\documentclass[conference]{IEEEtran}
\IEEEoverridecommandlockouts

\usepackage{cite}
\usepackage{amsmath,amssymb,amsfonts}
\usepackage{algorithmic}
\usepackage{graphicx}
\usepackage{textcomp}
\usepackage{xcolor}
\usepackage{url}
\usepackage{hyperref}
\usepackage{booktabs}
\usepackage{longtable}
\usepackage{array}

\newcommand{\citep}[1]{\cite{#1}}
\newcommand{\citet}[1]{\cite{#1}}

\def\BibTeX{{\rm B\kern-.05em{\sc i\kern-.025em b}\kern-.08em
    T\kern-.1667em\lower.7ex\hbox{E}\kern-.125emX}}

\newcounter{xaviercounter}

\newcounter{yanniscounter}

\begin{document}

\title{Echo State Transformer:\\Attention over Finite Memories}


\author{\IEEEauthorblockN{Yannis Bendi-Ouis}
\IEEEauthorblockA{\textit{Mnemosyne} \\
\textit{Centre Inria de l'Université de Bordeaux}\\
Bordeaux, France \\
yannis.bendi-ouis@inria.fr}
\and
\IEEEauthorblockN{Xavier Hinaut}
\IEEEauthorblockA{\textit{Mnemosyne} \\
\textit{Centre Inria de l'Université de Bordeaux}\\
Bordeaux, France \\
xavier.hinaut@inria.fr}
}

\maketitle

\begin{abstract}
While Large Language Models and their underlying Transformer architecture are remarkably efficient, they do not reflect how our brain processes and learns a diversity of cognitive tasks such as language, nor how it leverages working memory.
Furthermore, Transformers encounters a computational limitation: quadratic complexity growth with sequence length. 
Motivated by these limitations, we aim to design architectures that leverage efficient working memory dynamics to overcome standard computational barriers.
We introduce Echo State Transformers (EST), a hybrid architecture that resolves this challenge while demonstrating state of the art performance in classification and detection tasks. 
EST integrates the Transformer attention mechanisms with nodes from Reservoir Computing to create a fixed-size memory system. Drawing inspiration from Echo State Networks, our approach leverages several reservoirs (random recurrent networks) in parallel as a lightweight and efficient working memory.
These independent units possess distinct and learned internal dynamics with an adaptive leak rate, enabling them to dynamically adjust their own temporality.
By applying attention on those fixed number of units instead of input tokens, EST achieves linear complexity for the whole sequence, effectively breaking the quadratic scaling problem of standard Transformers. We evaluate ESTs on a recent timeseries benchmark: the Time Series Library, which comprises 69 tasks across five categories. Results show that ESTs ranks first overall in two of five categories, outperforming strong state-of-the-art baselines on classification and anomaly detection tasks, while remaining competitive on short-term forecasting. 
These results demonstrate that by shifting the attention mechanism from the entire input sequence to a fixed set of evolving memory units, it is possible to maintains high sensitivity to temporal events while achieving constant computational complexity per step.
\end{abstract}

\begin{IEEEkeywords}
Transformer, Reservoir Computing, Time Series
\end{IEEEkeywords}

\section{Introduction}
Transformers \citep{vaswani2017attention} have constituted a genuine revolution in the field of artificial intelligence, offering for the first time a scalable architecture capable of efficiently processing text while overcoming the inherent constraints of classical Recurrent Neural Networks (RNNs) and their costly training related to backpropagation through time \citep{werbos1990backpropagation}. This major innovation, comes with a significant drawback: a quadratic complexity that increases with the length of the sequence to be processed.
Although Transformers break free from  RNN formalism and their internal states 
(transmitting information from time $t$ to time $t+1$)
they still need to access previous information. To achieve this, rather than using only the current information, they mobilize the entire input sequence at each step of the processing, meaning a Transformer has no internal memory per se, but must re-process the entire sequence at each time step. This characteristic generates several major problems. The most obvious concerns the energy and computational costs that inexorably increase as the conversation 
with a LLM (implemented via a Transformer) lengthens.
A second problem lies in the fact that a Transformer, like a human being, eventually becomes overwhelmed when the sequence becomes particularly long \citep{modarressi2025nolima}. Moreover, having to retain the entire sequence is equivalent to possessing an \textit{infinite} working memory, which is considerably far from what is biologically plausible~\citep{baddeley1992working}. Indeed, when we humans read a book, we are incapable of memorizing all the words and punctuation elements that compose it. Instead, we have a \textit{finite} memory in which we \textit{compress} information. 
Biological systems show us that efficient ways to manage memory are possible, thus we take inspiration from Working Memory models~\citep{strock2020robust}.

\begin{figure}[t]
  \centering
  \includegraphics[width=\linewidth]{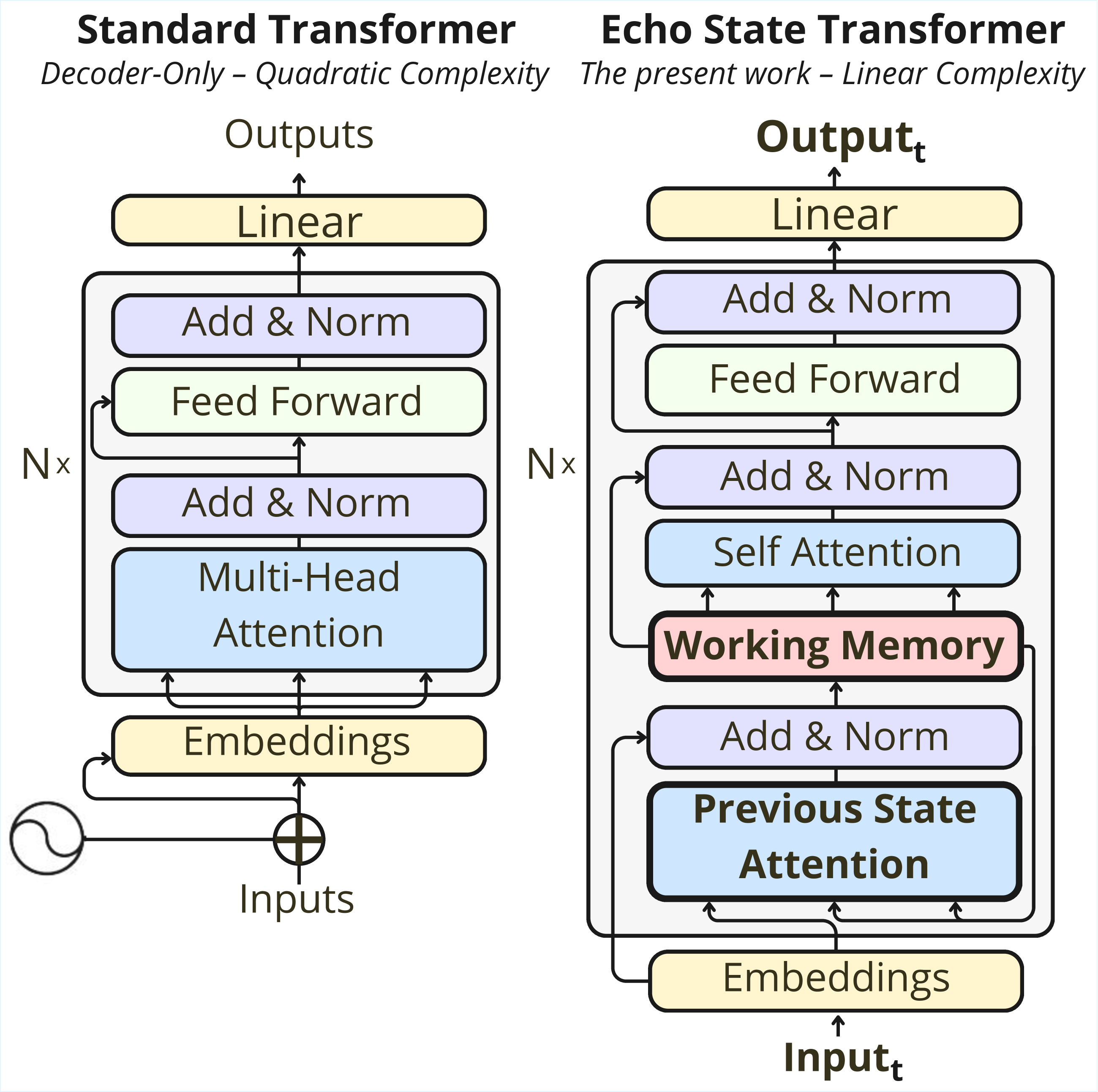}
  \caption{Standard Transformer (Decoder-Only) and Echo State Transformer (Ours) architectures. We add two blocks to Transformers: 
  a "Working Memory" (WM) block and a "Previous State Attention" (PSA) block. The "WM block which maintains a state across the sequence and the PSA block reads both memory and $Input_t$ to shape the individual input of each unit of the WM block.
  While standard Transformers process the entire sequence at once with quadratic complexity, our architecture operates sequentially like an RNN, processing one element at a time with constant computational cost per step, resulting in linear complexity with respect to sequence length.}
  \label{fig:archi_update}
  \vspace{-0.2cm}
\end{figure}

To overcome this limitation, we propose to add a Working Memory block to the standard Transformer architecture (Fig. \ref{fig:archi_update}) by taking inspiration from Reservoir Computing \citep{lukovsevivcius2009reservoir, yan2024emerging}. 
To achieves constant complexity per timestep (and thus linear complexity for the whole sequence), the core of our method 
is to exploit the attention computations of Transformers, not on the entire input sequence, but on a finite set of memory units. Each of these units has its own dynamics and must retain a specific part of the information received as input. Thus, since the number of memory units is determined at the initialization of the model and remains constant (unlike the number of tokens in the input sequence which continuously grows), the complexity of the attention computation also becomes constant.
For each timestep, each memory unit creates (i) its own Queries from the current input embedding at time $t$, (ii) its own Keys and Values based on both its own state and the states of other memory units. This mechanism allows each unit to independently extract from the collective memory the information it deems relevant and compare it to the model's input to shape its own input. Once the memories are updated, a second attention mechanism allows their content to be exploited to predict the output at time $t$.
%
%
Paper organisation: related works, architecture inspiration, model architecture, experiments, results, ablation study, FLOPs comparison, conclusion.

\section{Related works}

Numerous approaches have been explored to address the complexity of Transformers. Among them, many seek to modify, replace, or eliminate attention computation, particularly the \textit{softmax} operation which prevents recurrent formulation and requires the entire sequence to be reprocessed at each time step. Thus, \emph{Performers} \citep{choromanski2020rethinking} estimate the attention matrix with linear complexity using \emph{FAVOR+}. \emph{Attention Free Transformers} (AFT) \citep{zhai2021attention} eliminate self-attention by first combining Keys and Values with a set of positional biases before multiplying the result with Queries, achieving linear complexity. \emph{Reformer} \citep{kitaev2020reformer} replaces the attention product with locality-sensitive hashing offering linearithmic complexity $\mathcal{O}(n \log n)$. 
\emph{RWKV} \citep{peng2023rwkv} replaces attention and MLP blocks with two blocks allowing the mixing of temporal and spatial information, thus offering a linear solution in time and space. \emph{Retentive Network} (RetNet) \citep{sun2023retentive} replaces the attention softmax with an exponential decay applied to the product result of Queries and Keys, allowing three modes of usage including linear complexity inference. \emph{Mamba} \citep{gu2023mamba} proposes an innovative architecture based on selective \emph{State Space Models} (SSMs) and allows linear training and linear inference. \emph{TimesNet} \citep{wu2022timesnet} reshapes time series into 2D tensors over learned periods and applies lightweight convolutional inception blocks to capture temporal patterns, which leads to a linear complexity in sequence length.
\newpage
Other approaches 
that do not seek to modify the attention computation but rather redesign the representation of the input sequence. 
This is the case with \emph{TransformerFAM} \citep{hwang2024transformerfam} which introduces a feedback loop designed to dynamically create two tokens representing past information, thus creating an internal memory to process sequences of indefinite length. \emph{iTransformer} \citep{liu2023itransformer} inverts tokenization by treating each variate as a token instead of each time step, enabling attention to capture cross-variate dependencies while feed-forward networks learn temporal patterns. \emph{PatchTST} \citep{nie2022time} splits each univariate series into patches and applies channel-independent attention, enabling longer lookback windows and improved long-term forecasting. \emph{Large Memory Model} (LM2) \citep{kang2025lm2} proposes a memory module capable of interacting with input tokens and updating itself via gate mechanisms, similar to LSTMs \citep{hochreiter1997long}, to store and compress past information. 

\section{Architecture inspiration}
Our \emph{Echo State Transformer} architecture draws inspiration from two powerful paradigms in sequential processing: \emph{Transformers} and \emph{Reservoir Computing}. This section  introduces the key components that we leverage in our hybrid model.

\subsection{Transformers}

\begin{figure}[h]
  \centering
  \includegraphics[width=0.95\linewidth]{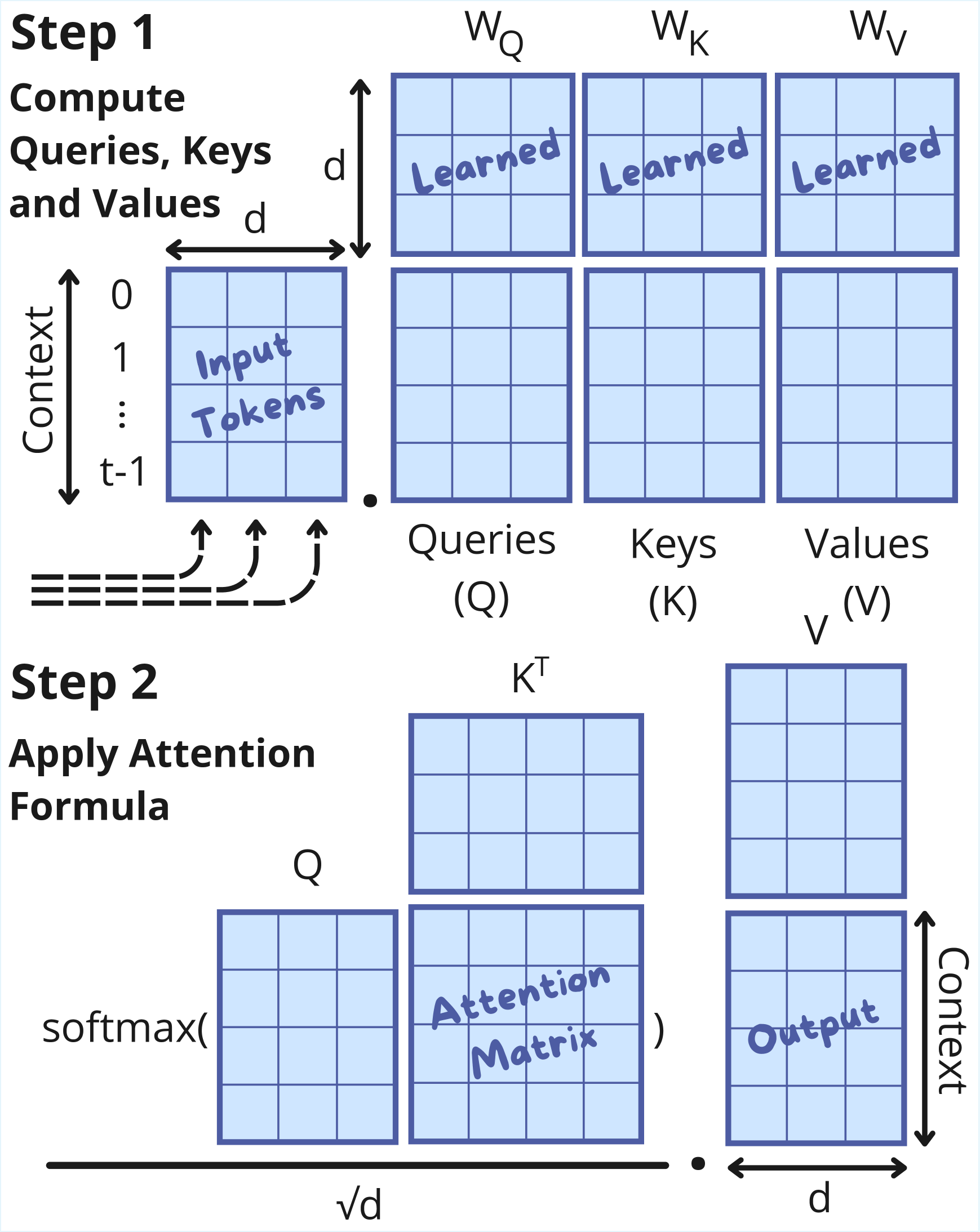}
  \caption{Attention computation in Transformers exemplified with matrix multiplication diagrams. 
  (top) Computation of Queries, Keys and Values. 
  (bottom) Application of the attention formula with the previous computed Queries, Keys and Values.}
  \label{fig:attention}
\end{figure}

Transformers revolutionized sequence processing by replacing recurrence with attention mechanisms, allowing direct interaction between any positions in a sequence. The core innovation of Transformers is the self-attention mechanism:
\begin{equation}
\text{Attention}(Q, K, V) = \text{softmax}\left(\frac{QK^T}{\sqrt{d_k}}\right)V
\end{equation}
where $Q$ (queries), $K$ (keys), and $V$ (values) are linear projections of the input. This mechanism computes a weighted sum of values, with weights determined by the compatibility between queries and keys.
Transformers process sequences through multiple layers, each containing a Self-Attention and Feed-Forward modules. Recent research suggests these feed-forward layers act as key-value memories \citep{geva2020transformer}, compatible with brain mechanisms assumptions 
\citep{gershman2025key}, storing knowledge acquired during training.
While Transformers excel at modeling complex dependencies and can process sequences in parallel, they face a quadratic complexity challenge with sequence length and lack an inherent mechanism for maintaining state across processing steps.

\subsection{Reservoir Computing}

\begin{figure}[h]
  \centering
  \includegraphics[width=0.9\linewidth]{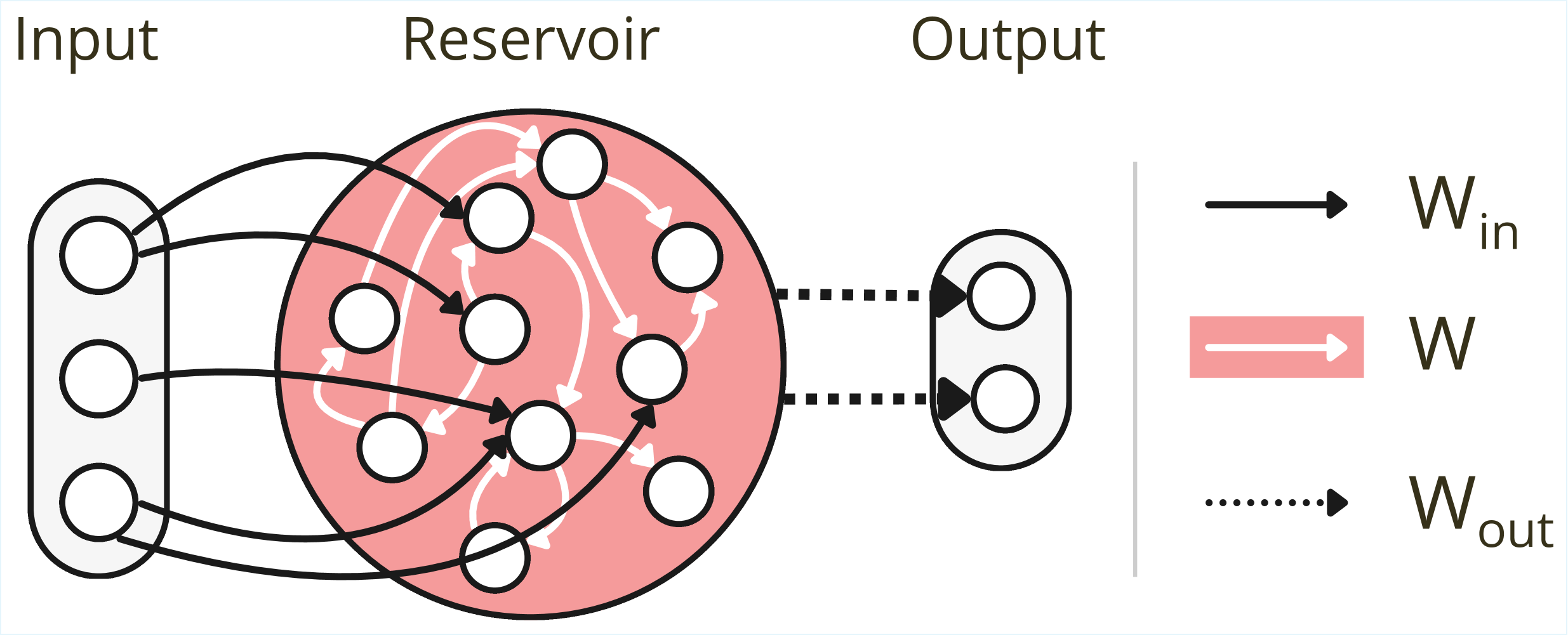}
  \caption{Echo State Network is composed of 3 matrix weights: $W_{in}$ treats the input, $W$ computes the state update and $W_{out}$ computes the output. Only $W_{out}$ is trained via linear or ridge regression.}
  \label{fig:esn}
  \vspace{-0.1cm}
\end{figure}

Reservoir Computing, particularly Echo State Networks (ESNs, see Fig. \ref{fig:esn}) \citep{jaeger2001echo, jaeger2004harnessing} offers an efficient approach to sequential processing through fixed random recurrent networks. The key dynamics of a reservoir are captured by:
\begin{equation}
\mathbf{s}_t = (1-\alpha).\mathbf{s}_{t-1} + \alpha.f(\mathbf{W}_{in}.\mathbf{u}_t + \mathbf{W}.\mathbf{s}_{t-1})
\end{equation}
where $\mathbf{s}_t$ is the reservoir state, $\mathbf{u}_t$ is the input, $\mathbf{W}_{in}$ and $\mathbf{W}$ are fixed random matrices, $f$ is a non-linear activation (typically $\tanh$), and $\alpha$ is the leak rate.
Two critical parameters govern reservoir behavior: the spectral radius --  defined as the largest absolute eigenvalue of $\mathbf{W}$ -- controls the Echo State Property \citep{yildiz2012re}, with values below 1 ensuring stability and values approaching 1 maximizing memory capacity while maintaining predictable behavior; complementing this, the leak rate ($\alpha$) regulates how quickly the reservoir state evolves \citep{jaeger2007optimization}, where lower values preserve previous states longer and higher values enhance the network's responsiveness to new inputs.
Reservoirs excel at maintaining temporal information with minimal parameter tuning and training data requirements \citep{lukovsevivcius2009reservoir}. By only training output weights, ESNs achieve remarkable efficiency while maintaining powerful temporal processing capabilities \citep{jaeger2001echo}. If properly configured, reservoirs can retain temporal information proportional to their size \citep{ceni2024edge}.

\section{Model architecture}

Our approach, the Echo State Transformer (EST)\footnote{Full implementation available on our anonymous public git repository:\\  \indent https://anonymous.4open.science/r/EchoStateTransformer/}, proposes a hybrid architecture inspired by standard Transformers and Reservoir Computing. Our main objective is to overcome the quadratic complexity inherent to Transformers. To do so, the core idea is to compute Queries, Keys and Values on a finite set of memory states (i.e. one state per memory units) instead of an "infinite" sequence of tokens. That way, the attention matrix remains the same size at each timestep (i.e $M \times M$, where $M$ is the number of memory units), making each step constant in complexity.
Another innovation proposed is the use of an adaptive leak rate. It allows better control of information retention over time, compensating for the gradual dissipation of information. 
In preliminary works \citep{bendi2024recurrent} we observed that each memory unit tends to stabilize around a characteristic leak rate value, oscillating dynamically in response to input variations. This emergent property allows each unit to maintain a consistent temporal profile while still adapting to contextual needs. 
The EST mainly consists of three distinct blocks (Fig. \ref{fig:archi_update}): an input block, an output block, and four sub-blocks that can be stacked multiple times together to form several layers. These four sub-blocks are: an attention block on the previous state, a working memory block, a self-attention block, a feed-forward block.

\subsection{Input Block}
The model transforms a raw input vector $\mathbf{x}_t$ at each time step $t$ into an embedded vector $\mathbf{e}_t \in \mathbb{R}^{1 \times D}$, where $D$ denotes the model dimension:
\begin{equation}
    \mathbf{e}_t = \text{Embed}(\mathbf{x}_t)
\end{equation}

\subsection{Previous State Attention Block}

\begin{figure}[h]
    \centering
    \includegraphics[width=0.99\linewidth]{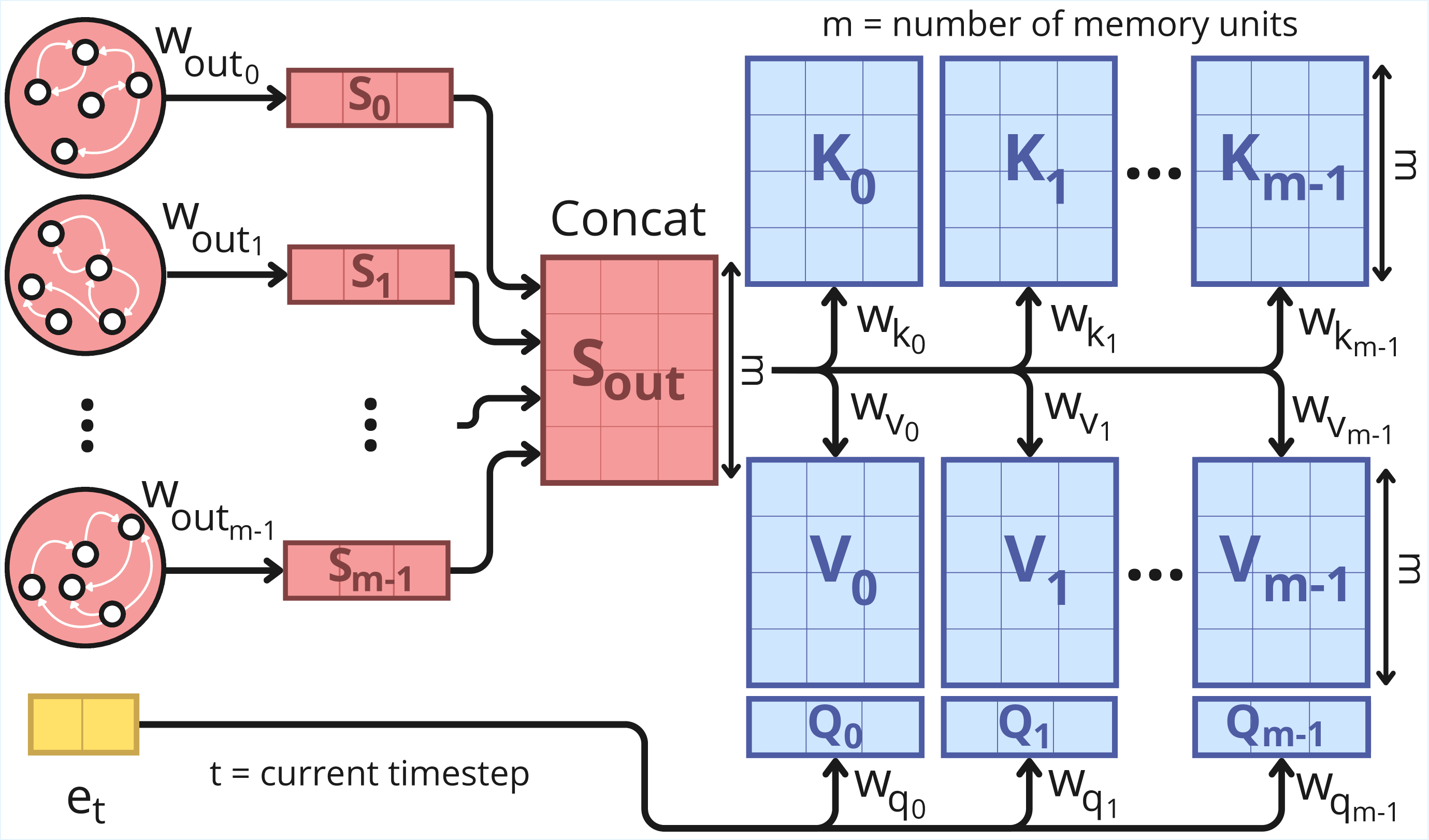}
    \caption{
    How the Previous State Attention block produces, for each time step, Keys and Values from all memory units ($S_{out}$), and Queries from the embedding ($e_t$). Then, for each memory unit, a unique triplet (Q,K,V) is produced. These triplets are then used to compute distinct attention products to create input vector $u(t)$ for each memory units a unique.}
    \label{fig:attention_previous_state}
\end{figure}

To allow each of the $M$ memory units to retrieve specific context, the model performs attention between the current input and the previous memory state. As shown in Fig. \ref{fig:attention_previous_state} Queries are derived from the input $\mathbf{e}_t$, while Keys and Values are projected from the previous memory output $\mathbf{s}_{t-1} \in \mathbb{R}^{M \times D}$ (initialized with zeros):
\begin{equation}
    \mathbf{Q}_{psa} = \mathbf{e}_t \mathbf{W}_{Q_{psa}} 
\end{equation}
\begin{equation}
    \mathbf{K}_{psa} = \mathbf{s}_{t-1} \mathbf{W}_{K_{psa}} \quad \quad \mathbf{V}_{psa} = \mathbf{s}_{t-1} \mathbf{W}_{V_{psa}}
\end{equation}
where $\mathbf{W}_{Q_{psa}}, \mathbf{W}_{K_{psa}}, \mathbf{W}_{V_{psa}} \in \mathbb{R}^{M \times D \times D}$ are learnable matrices. The attention mechanism weights the input relevance for each unit, producing an update vector $\mathbf{u}_t \in \mathbb{R}^{M \times 1 \times D}$ via a residual connection and RMSNorm:
\begin{equation}
    \mathbf{a}_{psa} = \text{Softmax}\left( \frac{\mathbf{Q}_{psa} \mathbf{K}_{psa}^\top}{\sqrt{D}} \right) \mathbf{V}_{psa}
\end{equation}
\begin{equation}
    \mathbf{u}_t = \text{RMSNorm}\left( \mathbf{e}_t + \mathbf{a}_{psa} \right)
\end{equation}

\subsection{Working Memory Block}

\begin{figure}[h]
  \centering
  \includegraphics[width=0.95\linewidth]{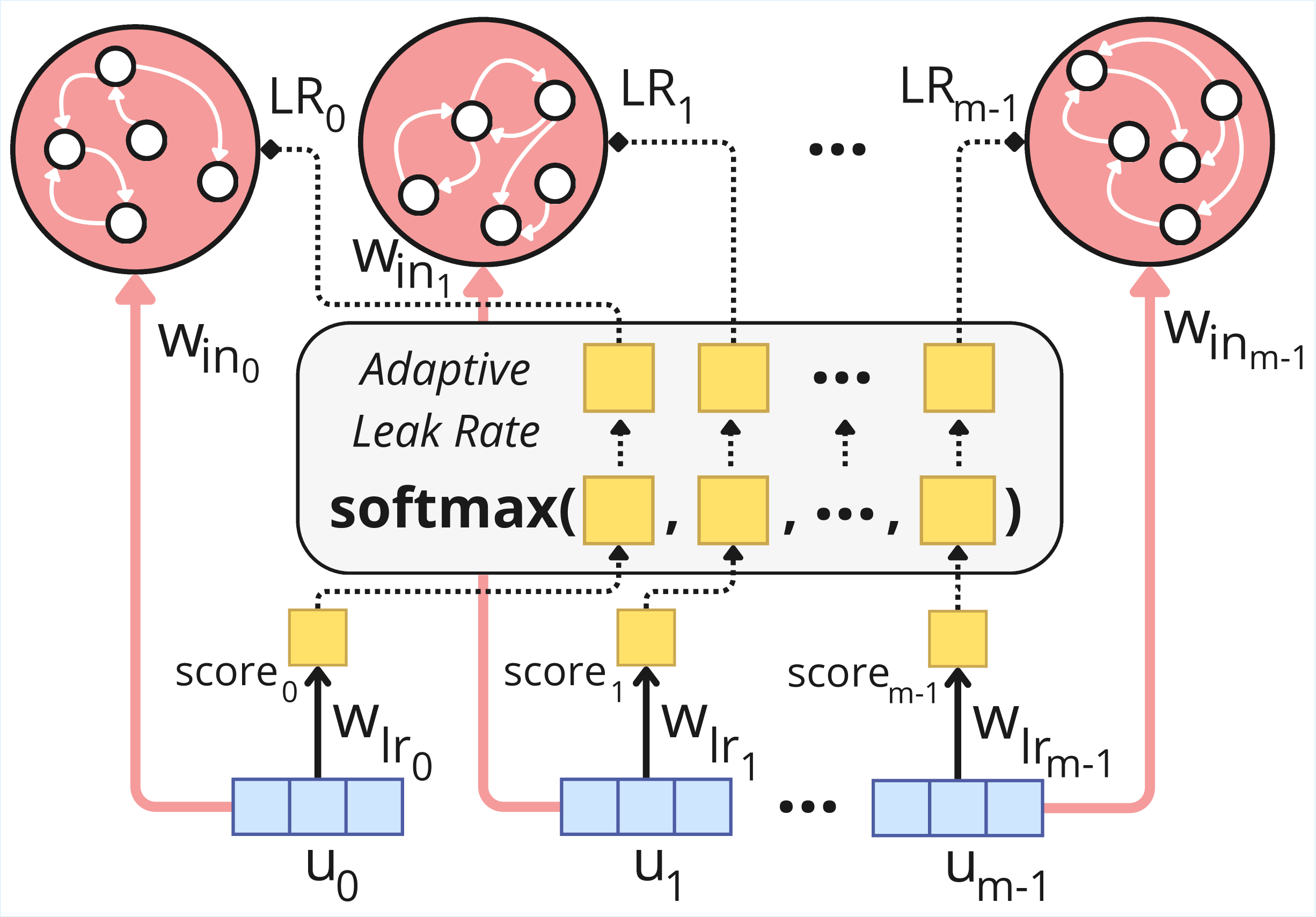}
  \caption{
  Mechanisms behind the Working Memory block including the adaptive leak rate. Each memory unit compute a score $score_i$ from its input vector $u_i$. Then, a softmax is applied on all scores to compute the leak rate $LR_i$ for each unit. This competitive mechanism 
  allows certain units to maintain fixed information over time by assigning near-zero leak rates. }
  \label{fig:working_memory}
  \vspace{-0.15cm}
\end{figure}

The core memory relies on Reservoir Computing dynamics with an adaptive leak rate mechanism and learned spectral radii. First, a leak rate $\boldsymbol{\alpha}_{t} \in [0, 1]^{M}$ is computed dynamically for each unit (see Fig. \ref{fig:working_memory}) using a learnable projection $\mathbf{W}_{\alpha} \in \mathbb{R}^{M \times D \times 1}$ and temperature $\tau$, determining the balance between retaining previous information (value near 0) and accepting new input (value near 1):
\begin{equation}
    \boldsymbol{\alpha}_{t} = \text{Softmax}\left( \frac{\mathbf{u}_{t} \mathbf{W}_{\alpha}}{\tau} \right)
\end{equation}
The internal reservoir state $\mathbf{h}_t \in \mathbb{R}^{M \times 1 \times R}$ (with dimension $R$) is updated using sparse input connectivity $\mathbf{W}_{in} \in \mathbb{R}^{M \times D \times R}$ and sparse recurrent weights $\mathbf{W}_{res} \in \mathbb{R}^{M \times R \times R}$:
\begin{equation}
    \tilde{\mathbf{h}}_t = \tanh\left( \mathbf{u}_{t} \mathbf{W}_{in} + \mathbf{h}_{t-1} \mathbf{W}_{res} + \mathbf{b} \right)
\end{equation}
\begin{equation}
    \mathbf{h}_t = (1 - \boldsymbol{\alpha}_{t}) \odot \mathbf{h}_{t-1} + \boldsymbol{\alpha}_{t} \odot \tilde{\mathbf{h}}_t
\end{equation}

\vspace{-0.45cm}

Finally, the state is projected back to the model dimension via the readout matrix $\mathbf{W}_{out} \in \mathbb{R}^{M \times R \times D}$ to produce $\mathbf{s}_t \in \mathbb{R}^{M \times 1 \times D}$ that we will use as $\mathbb{R}^{M \times D}$ later:
\begin{equation}
    \mathbf{s}_t = \mathbf{h}_t \mathbf{W}_{out}
\end{equation}

\subsection{Self Attention Block}
To model dependencies between the memory units themselves, self-attention is applied to the updated state $\mathbf{s}_t$. Treating memory units as tokens, we compute queries, keys, and values using weights $\mathbf{W}_{Q_{sa}}, \mathbf{W}_{K_{sa}}, \mathbf{W}_{V_{sa}} \in \mathbb{R}^{D \times D}$:
\begin{equation}
    \mathbf{Q}_{sa} = \mathbf{s}_t \mathbf{W}_{Q_{sa}}, \quad \mathbf{K}_{sa} = \mathbf{s}_t \mathbf{W}_{K_{sa}}, \quad \mathbf{V}_{sa} = \mathbf{s}_t \mathbf{W}_{V_{sa}}
\end{equation}
The output $\mathbf{v}_t \in \mathbb{R}^{M \times D}$ incorporates a residual connection and normalization:
\begin{equation}
    \mathbf{a}_{sa} = \text{Softmax}\left( \frac{\mathbf{Q}_{sa} \mathbf{K}_{sa}^\top}{\sqrt{D}} \right) \mathbf{V}_{sa}
\end{equation}
\begin{equation}
    \mathbf{v}_t = \text{RMSNorm}\left( \mathbf{s}_t + \mathbf{a}_{sa} \right)
\end{equation}

\subsection{Feed Forward Block}
The attention result $\mathbf{v}_t$ is fused by concatenation and projected by $\mathbf{W}_{reduce} \in \mathbb{R}^{(M \cdot D) \times D}$ into a latent vector $\mathbf{z}_t$. A standard Feed-Forward network integrates learned knowledge \citep{geva2020transformer} to produce the layer output $\mathbf{y}_t \in \mathbb{R}^{D}$:
\begin{equation}
    \mathbf{z}_t = \text{Concat}(\mathbf{v}_{t,1}, \dots, \mathbf{v}_{t,M}) \mathbf{W}_{reduce}
\end{equation}
\begin{equation}
    \mathbf{h}_{ff} = \text{GELU}(\mathbf{z}_t \mathbf{W}_{1}) \mathbf{W}_{2}
\end{equation}
\begin{equation}
    \mathbf{y}_t = \text{RMSNorm}\left( \mathbf{z}_t + \mathbf{h}_{ff} \right)
\end{equation}

\subsection{Output Block}
Depending on the task, the final prediction $\hat{\mathbf{Y}}$ is obtained via a projection $\mathbf{W}_{proj}$ applied either to the full sequence $\mathbf{Y}_{seq}$ (e.g., classification) or the current step $\mathbf{Y}_{t}$ (e.g., anom. detec.):
\begin{equation}
    \hat{\mathbf{Y}} = \mathbf{Y}_{seq} \mathbf{W}_{proj} + \mathbf{b}_{proj} \quad \text{or} \quad \hat{\mathbf{Y}}_t = \mathbf{Y}_{t} \mathbf{W}_{proj} + \mathbf{b}_{proj}
\end{equation}

\section{Experiments}

\subsection{Benchmark framework}
We evaluate our model on the Time Series Library (TSL) benchmark \citep{wang2024deep}, which provides a unified evaluation protocol for deep time series models. TSL encompasses five task categories, each with its own evaluation metric and multiple individual tasks, for a total of 69 tasks: Anomaly Detection (5 tasks, evaluated using F1-score), Classification (10 tasks, evaluated using Accuracy), Imputation (12 tasks, evaluated using Mean Squared Error), Long-term Forecasting (36 tasks, evaluated using Mean Squared Error), and Short-term Forecasting (6 tasks, evaluated using Overall Weighted Average).

\subsection{Evaluation setup}
We compare the proposed Echo State Transformer (EST) with state-of-the-art baselines present in \citet{wang2024deep}. All experiments strictly follow the training protocols defined in TSL to ensure fair comparison, including optimization strategies, batch construction, and evaluation methodology. Following the TSL methodology, we tested ten different configurations\footnote{All codes used for our experiments (including all configurations) are available on our anonymous public git repository: https://anonymous.4open.science/r/EchoStateTransformer/} (e.g. number of layers, memory units, neurons, connectivity) and retained the best-performing one for each task.

\section{Results}

We evaluate EST across 69 tasks grouped in 5 time series problem categories, comparing it against baselines (including state-of-the-art models) whose results in Table~\ref{table:results} are taken from \citep{wang2024deep}. EST ranks $1^{st}$ overall in 2 of the 5 task categories, demonstrating competitive performance on a diverse set of tasks.

\begin{table}[htbp]
\caption{Comparative results on TSL Benchmark. \\ Best in bold, second best underlined.}
\label{table:results}
\centering
\small 
\setlength{\tabcolsep}{1pt} 
\begin{tabular}{lccccc}
\toprule
\textbf{Modèle} & \textbf{Long-term} & \textbf{Short-term} & \textbf{Imput.} & \textbf{Classif.} & \textbf{Anom. Det.} \\
 & (MSE $\downarrow$) & (OWA $\downarrow$) & (MSE $\downarrow$) & (Acc. $\uparrow$) & (F1 $\uparrow$) \\
\midrule
Autoformer   & 0.436 & 0.992 & 0.062 & 71.07 & 84.26 \\
Crossformer  & 0.559 & 1.075 & 0.077 & 73.17 & 83.45 \\
DLinear      & 0.414 & 1.130 & 0.119 & 70.71 & 82.46 \\
FEDformer    & 0.403 & 0.945 & 0.112 & 70.67 & 84.97 \\
iTransformer & \textbf{0.342} & 1.030 & 0.117 & 70.06 & 81.39 \\
Mamba        & 0.429 & 0.922 & 0.069 & 69.87 & 83.89 \\
N-BEATS      & \underline{0.371} & 0.890 & 0.190 & 68.96 & 81.93 \\
PatchTST     & 0.373 & \textbf{0.871} & 0.065 & 72.50 & 82.79 \\
SCINet       & 0.589 & --    & 0.059 & 71.71 & 84.74 \\
Stationary   & 0.427 & 1.000 & \underline{0.057} & 72.66 & 82.10 \\
TiDE         & 0.459 & --    & 0.118 & --    & --    \\
TimesNet     & 0.381 & \underline{0.888} & \textbf{0.050} & \underline{73.67} & \underline{85.24} \\
\midrule
\textbf{EST (Ours)} & 0.774 & 0.929 & 0.134 & \textbf{74.08} & \textbf{85.25} \\
\bottomrule
\end{tabular}
\end{table}

\subsection{EST achieves SOTA performance on classification and anomaly detection}

EST seems to excel at identifying minor variations within complex data streams. 
In classification, EST achieves 74.08\% average accuracy, ranking $1^{st}$ overall among all evaluated models and outperforming leading approaches including TimesNet (73.67\%), Crossformer (73.17\%), and PatchTST (72.50\%). 
Similarly for anomaly detection, EST secures the top position with an F1-score of 85.25\% in par with TimesNet (85.24\%) and surpassing FEDformer (84.97\%). These results demonstrate EST's capability in temporal pattern analysis, excelling both at categorical classification and at detecting deviations from normal behavior.

\subsection{Competitive performance on short-range forecasting}

While not leading the short-range, EST (0.929) performs competitively with established methods like Mamba (0.922) and substantially outperforms several baselines including iTransformer (1.030) and Crossformer (1.075). This suggests EST effectively captures local temporal dependencies over modest prediction horizons.

\subsection{Architecture insights from optimal configurations}

Analysis of the best-performing configurations reveals interesting architectural patterns across task types. For classification and anomaly detection tasks EST performs best with deeper architectures of 2-4 layers, suggesting that hierarchical feature extraction benefits pattern recognition. In contrast, imputation and short-term forecasting predominantly select shallow 1-layer configurations, indicating these tasks benefit from very short-term temporal modeling.
Additionally, as expected, certain tasks require greater memory capacity than others. Short-term forecasting tasks consistently favor balanced configurations where memory dimensions closely match attention computation dimensions (e.g., mem64-dim64, mem32-dim32), suggesting that short-term information retention benefits from symmetric representational capacity. In contrast, tasks requiring long-term information retention exhibit a clear preference for substantially larger memory dimensions relative to model dimensions (e.g., mem128-dim64, mem512-dim128). This pattern is consistent with the expectation that larger reservoir enable longer information retention, providing the extended temporal context necessary for these tasks.
More broadly, optimal ESTs seems to be usable in various scales, with a minimum of 4 memory units up to 16 units (in our tests) depending on task complexity. Configurations with only 2 memory units are effective only in rare cases, 
typically for simple patterns or when coupled with high-dimensional memory spaces (e.g., 512mem). 
These configuration patterns demonstrate EST's architectural flexibility: memory dimensions can be scaled to handle varying temporal horizons, while layer depth and model size can be adjusted to match the required feature extraction complexity.

\subsection{Limitations on reconstruction tasks}

EST lags behind state-of-the-art models in reconstruction tasks (Imputation, Long-term Forecasting), reflecting that reservoir dynamics capture global system behaviors rather than precise pointwise values \citep{lu2018attractor} \citep{cuchiero2022discrete}. Additionally, the optimization process remains constrained by BPTT, which limits the effective modeling of long-range dependencies.

\section{Ablation study}
To validate our architecture, we assessed the impact of the Previous State Attention (PSA) and Adaptive Leak Rate (ALR) mechanisms (see Table \ref{tab:ablation}). Results indicate that ALR is critical for temporal dynamics, particularly in forecasting tasks where its removal causes a severe performance drop. Conversely, PSA is essential for pattern recognition, significantly boosting performance in Classification and Anomaly Detection but seems not critical for short-term forecast. Overall, the full EST model achieves the best trade-off across all tasks, demonstrating the importance of both components.

\begin{table}[h]
\caption{Ablation Study: \\Importance of Previous State Attention (PSA) and Adaptive Leak Rate (ALR) on performances. \\Best in bold, second best underlined.
}
\label{tab:ablation}
\centering
\small 
\setlength{\tabcolsep}{1pt} 
\begin{tabular}{lccccc}
\toprule
\textbf{Variante} & \textbf{Long-term} & \textbf{Short-term} & \textbf{Imput.} & \textbf{Classif.} & \textbf{Anom. Det.} \\
 & (MSE $\downarrow$) & (OWA $\downarrow$) & (MSE $\downarrow$) & (Acc. $\uparrow$) & (F1 $\uparrow$) \\
\midrule
\textbf{EST (Full)} & \textbf{0.774} & \underline{0.929} & \textbf{0.134} & \textbf{74.08} & \textbf{85.25} \\
w/o PSA             & \underline{0.838} & \textbf{0.928} & 0.145 & 70.57 & 80.58 \\
w/o ALR     & 1.113 & 1.079 & \underline{0.135} & \underline{71.57} & \underline{82.05} \\
\bottomrule
\end{tabular}
\end{table}

\section{FLOPs comparison}

\begin{figure}[h]
  \centering
  \includegraphics[width=\linewidth]{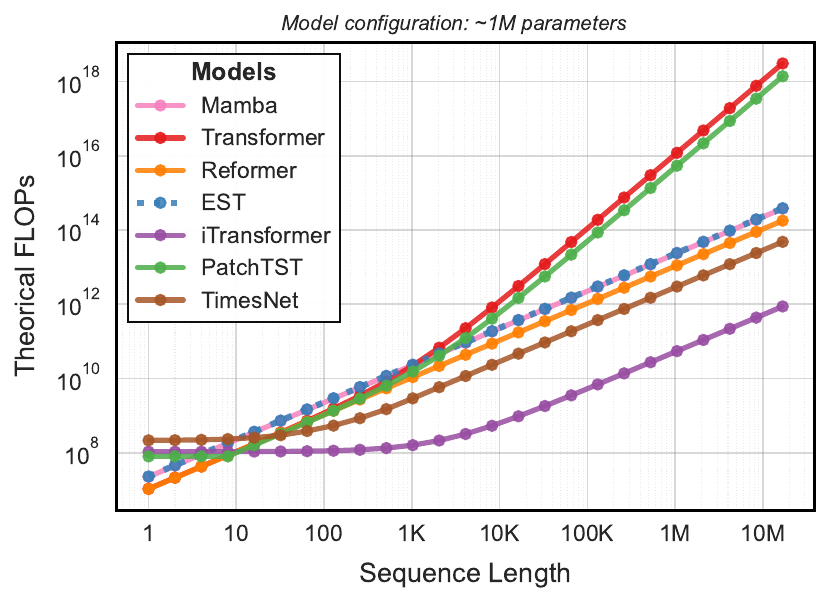}
  \caption{Theoretical FLOPs vs. sequence length for models normalized to $\approx$ 1M parameters. EST (in blue) and Mamba (in pink) lines are superimposed.}
  \label{fig:flops}
\end{figure}

We compute theoretical FLOPs per forward pass by summing all additions and multiplications across shared components (embeddings, normalization, linear projections, feed-forward layers, activations, softmax) and model-specific blocks (e.g., attention patterns, SSM updates, convolutions). Non-arithmetic operations such as dropout, reshapes, and permutations are excluded. To isolate the impact of sequence length scaling, all models in Fig.~\ref{fig:flops} share similar depth and width configurations, resulting in $\approx$~1M parameters.

Complexity analysis reveals distinct scaling behaviors as sequence length $L$ increases. Transformer and PatchTST exhibit quadratic scaling $\mathcal{O}(L^2)$ due to standard self-attention. This quadratic growth dominates computational costs at longer sequences. In contrast, several architectures achieve more favorable scaling properties. Mamba maintains linear complexity $\mathcal{O}(L)$, positioning slightly above Reformer's $\mathcal{O}(L \log L)$, which behaves linearly in practice, due to the logarithmic term that contributes to relatively few operations compared to the linear term. TimesNet and iTransformer display hybrid scaling: TimesNet transitions to linear growth after initial FFT costs (below $\approx 100$ tokens), while iTransformer remains nearly constant up to $\approx 500$ tokens before linear scaling emerges, driven by $\mathcal{O}(N^2)$ attention over fixed variables rather than time steps. Similarly, our proposed EST model achieves $\mathcal{O}(L)$ complexity by attending to fixed memory units instead of whole input sequence, closely tracking Mamba's efficiency.

\section{Conclusion}

We introduced the Echo State Transformer (EST), a recurrent architecture combining reservoirs and attention mechanisms with adaptive leak rate. On the large TSL benchmark, EST achieves state-of-the-art performance in classification and anomaly detection (ranking 1st overall), while having mixed performance in tasks demanding precise pointwise reconstruction like imputation. 
Crucially, this performance validates our architectural shift: by moving the attention mechanism from the entire input sequence to a fixed set of evolving memory units, the EST maintains high sensitivity to temporal events while achieving constant computational complexity per step. Future work will focus on removing BPTT for full sequence parallelization, requiring the removal of recurrent connection and the linearization of reservoir.

\section*{Acknowledgment}
Plafrim and Jean-Zay as HPC clusters.
Qwen, Gemini, ChatGPT and Claude for language editing and coding support.

\bibliography{refs}
\bibliographystyle{IEEEtran}

\newpage
\onecolumn
\appendices

\end{document}